\documentclass{article}

\PassOptionsToPackage{numbers, compress}{natbib}
\usepackage{graphicx} 
\usepackage{amsmath} 
\usepackage{algorithm}
\usepackage{comment}
\usepackage{ragged2e}

\usepackage{algpseudocode}
\usepackage[preprint]{neurips_2025}

\usepackage[utf8]{inputenc} 
\usepackage[T1]{fontenc}    
\usepackage{hyperref}       
\usepackage{url}            
\usepackage{booktabs}       
\usepackage{amsfonts}       
\usepackage{nicefrac}       
\usepackage{microtype}      
\usepackage{xcolor}         
\usepackage{wrapfig}

\newtheorem{hypt}{Conjecture}[section]

\title{Emergent Specialization: Rare Token Neurons in Language Models}

\author{
  Jing Liu \\
  ENS, Université PSL, EHESS, CNRS \\
  Paris, France \\
  \texttt{jing.liu@psl.eu} \\
  \And
  Haozheng Wang\\
  $^1$DT Master Carbon\\
  $^2$Independant Researcher\\
  Paris, France\\
  \texttt{haozheng.wang@ensea.fr} \\
  \And
  Yueheng Li\\
  $^1$Sorbonne Université, École Normale Supérieure, CNRS, Laboratoire de
Physique (LPENS)\\
  $^2$Laboratoire de Physique de l’École Normale Supérieure, ENS,\\
  Université PSL, CNRS, Sorbonne Université, Université Paris Cité\\
  Paris, France\\
  \texttt{yueheng.li@ens.fr} \\
}

\begin{document}

\maketitle

\begin{abstract}

Large language models struggle with representing and generating rare tokens despite their importance in specialized domains. In this study, we identify neuron structures with exceptionally strong influence on language model's prediction of rare tokens, termed as \textit{rare token neurons}, and investigate the mechanism for their emergence and behavior. These neurons exhibit a characteristic three-phase organization (plateau, power-law, and rapid decay) that emerges dynamically during training, evolving from a homogeneous initial state to a functionally differentiated architecture. In the activation space, rare token neurons form a coordinated subnetwork that selectively co-activates while avoiding co-activation with other neurons. This functional specialization potentially correlates with the development of heavy-tailed weight distributions, suggesting a statistical mechanical basis for emergent specialization.

\end{abstract}

\section{Introduction}

Large language models (LLMs) have demonstrated remarkable capabilities in learning statistical patterns of human language. However, these models consistently struggle with representing and generating rare tokens—words or phrases that appear infrequently in training data \citep{kandpal2023large, zhang2025systematic, mallen2023not}. This poses significant challenges for both basic language modeling and specialized domain applications, where rare but critical information often resides in the long tail of the distribution.

This challenge stems from the power-law distributions inherent in natural language \citep{zipf1949human, wyllys1981empirical}, with a significant portion of linguistic phenomena appearing with extremely low frequency\citep{gpt3, hoffmann2022training}. Recent work has highlighted how this challenge can lead to collapse when models are trained on synthetic data that either truncates or narrows the tail of the distribution \citep{dohmatob2024tale, hataya2023will, bohacek2025nepotistically}. While several external methods have been proposed to address this limitation—such as retrieval-augmented generation \citep{lewis2020retrieval}, in-context learning \citep{dong2022survey}, and non-parametric memory mechanisms \citep{borgeaud2022improving}—the fundamental question remains: do LLMs develop internal mechanisms specialized for processing rare tokens during pre-training?

This question draws inspirations from human language acquisition, where children demonstrate remarkable "fast mapping" abilities—learning new words after minimal exposure—from as young as 12 months of age \citep{carey1978acquiring, markson1997children}. Cognitive neuroscience explains this capability through the Complementary Learning Systems (CLS) theory \citep{mcclelland1995there, o2014complementary}, which posits that the brain employs two distinct neural systems: a neocortical system for gradual learning of distributed representations, and a hippocampal system specialized for rapid encoding of specific experiences, including rare events \citep{kumaran2016learning, schapiro2017complementary}. This biological specialization enables humans to effectively learn from both statistical regularities and singular experiences.

Recent advances in mechanistic interpretability have developed various techniques for analyzing individual neurons within transformer models. While studies have revealed that neurons encode interpretable features ranging from syntactic relationships \citep{manning2020emergent, finlayson2021causal} to semantic concepts \citep{gurnee2023finding, bricken2023monosemanticity}, most work has focused on common patterns rather than specialized mechanisms for rare events. Notable exceptions include work \citet{stolfo2024confidence} discovering neurons that modulate token logits proportionally to frequency. In this study, we focus on decoder-only Transformer-based models and extend their work to focus on rare tokens and investigate how individual neurons in the final MLP layer of transformer-based language models spontaneously specialize in processing these tokens during training.

Our analysis reveal three key findings: (i) LLMs develop dedicated "rare token neurons" that disproportionately impact the prediction of infrequent tokens; (ii) These specialized neurons emerge through distinct phase transitions during training, evolving from a homogeneous initial state to a functionally differentiated architecture; (iii) The emergence of specialized neuron groups correlates with the development of heavy-tailed weight distributions, suggesting a statistical mechanical basis for functional specialization.

\section{Background}

\subsection{Transformer architecture}   
In this study, we focus on the Multi-Layer Perceptron (MLP) sublayers. Given a normalized hidden state $x \in \mathbb{R}^{d_{\text{model}}}$ from the residual stream, the MLP transformation is defined as:
\begin{equation}
\text{MLP}(x) = W_{\text{out}} \phi(W_{\text{in}} x + b_{\text{in}}) + b_{\text{out}},
\end{equation}
where $W_{\text{in}} \in \mathbb{R}^{d_{\text{mlp}} \times d_{\text{model}}}$ and $W_{\text{out}} \in \mathbb{R}^{d_{\text{model}} \times d_{\text{mlp}}}$ are learned weight matrices, and $b_{\text{in}}, b_{\text{out}}$ are biases. The nonlinearity $\phi$ is typically a GeLU activation. We refer to individual entries in the hidden activation vector $\phi(W_{\text{in}} x + b_{\text{in}})$ as \emph{neurons}, indexed by their layer and position (e.g., \texttt{<layer>.<index>}). The activations $n$ represent post-activation values of these neurons. We selected the last layer as it directly projects into the unembedding matrix that produces token probabilities, which creates a computational bottleneck where feature integration must occur \citep{wei2022emergent}.

\subsection{Heavy-Tailed Self-Regularization (HT-SR) Theory} \label{sec:Heavy-Tailed Self-Regularization (HT-SR) Theory}

Heavy-Tailed Self-Regularization (hereafter \textit{HT-SR}) theory offers a spectral lens on neural network generalization\citep{martin2019traditional, martin2021implicit,lu2024alphapruning, couillet2022random}. Specifically, consider a neural network with $L$ layers, let $W_i$ denote a weight matrix extracted from the $i$-th layer, where $W_i \in \mathbb{R}^{m \times n}$ and $m \geq n$. We define the correlation matrix associated with $W_i$ as:
\[
X_i := W_i^\top W_i \in \mathbb{R}^{n \times n},
\]
which is a symmetric, positive semi-definite matrix. The empirical spectral distribution (ESD) of $X_i$ is defined as:
\[
\mu_{X_i} := \frac{1}{n} \sum_{j=1}^{n} \delta_{\lambda_j(X_i)},
\]
where $\lambda_1(X_i) \leq \cdots \leq \lambda_n(X_i)$ are the eigenvalues of $X_i$, and $\delta$ is the Dirac delta function. The ESD $\mu_{X_i}$ represents a probability distribution over the eigenvalues of the weight correlation matrix, characterizing its spectral geometry.

HT-SR theory proposes that successful neural network training exhibits heavy-tailed spectral behavior in the ESDs of certain weight matrices, due to self-organization toward a critical regime between order and chaos. Such heavy-tailed behavior is captured by various estimators,  and particularly informative among which is the power-law (PL) exponent $\alpha_{\text{Hill}}$, who estimates the tail-heaviness of the eigenvalue distribution. Low values of $\alpha_{\text{Hill}}$ (typically $\alpha < 2$) indicate heavy-tailed behavior, often interpreted as signs of functional specialization and self-organized criticality \citep{yang2023test}. A formal definition of $\alpha_{\text{Hill}}$ and the associated estimation procedure is provided in Section~\ref{sec:Weight Eigenspectrum}.

\section{Rare Token neuron analysis framework}   

\subsection{Rare Token Neuron Identification}
Inspired by prior work on confidence-regulation neurons \citep{stolfo2024confidence}, we hypothesize that certain neurons in language models functionally specialize for modulating token-level probabilities—particularly for \textbf{\textit{rare}} tokens that appear infrequently in the training corpus. From a theoretical perspective, such specialization aligns with principles of sparse coding \citep{olshausen1997sparse} and the information bottleneck framework \citep{tishby2000information}, where neural capacity is selectively allocated to maximize efficiency under uncertainty.

\paragraph{Ablation methodology}

To investigate the functional specialization hypothesis, we perform targeted ablation experiments on multiple language models, such as the Pythia model families \citep{biderman2023pythia}, with intermediate checkpoints and training set available (The Pile \citep{gao2020pile}). Following the intervention approach of \citet{stolfo2024confidence}, we assess neuron-level influence by performing **mean ablation** experiments, that is, fixing a specific neuron's activation to its mean value over a reference dataset. Formally, let \( i \in \{1, 2, \dots, d_{\text{mlp}}\} \) index a neuron in the MLP layer, and let \( n_i \in \mathbb{R} \) denote its activation. For a given input \( x \in \mathcal{X} \), let \( x \) represent the final hidden state (i.e., the output of the last transformer block). The mean-ablated hidden state \( \tilde{x}^{(i)} \) is then given by:

\begin{equation}
\tilde{x}^{(i)} = x + ( \bar{n}_i - n_i ) w^{(i)}_{\text{out}},
\end{equation}

where \( \bar{n}_i \) is the mean activation of neuron \( i \) across a reference subset of inputs, and \( w^{(i)}_{\text{out}} \) is the corresponding output weight vector.

\paragraph{Quantifying neuron influence}\label{paragraph:delta loss def}
To quantify the influence of each neuron \( i \), we compute the \textit{Neuron Effect} metric, defined as the expected absolute change in token-level loss upon ablation:
\begin{equation}
\Delta\text{loss}(i) = \mathbb{E}_{x \sim \mathcal{D}} \left| \mathcal{L}(\text{LM}(x), x) - \mathcal{L}(\text{LM}(\tilde{x}^{(i)}), x) \right|,
\end{equation}
where \( \text{LM}(x) \) denotes the model's output after applying LayerNorm and decoding, and \( \mathcal{L} \) represents the token-level cross-entropy loss.

\paragraph{Experimental setup}
For each neuron, we measure the effect by setting its activation value to its mean computed across a dataset of 25,088 tokens sampled from the C4 Corpus \citep{raffel2020exploring}. Given our focus on rare tokens, we implement a two-stage filtering process: at stage one, we retain tokens below the 50th percentile in the unigram frequency distribution of the training set; then at stage two, we restrict our analysis to valid, correctly spelled English words\footnote{Token was filtered with pyspellchecker library: \url{https://pypi.org/project/pyspellchecker/}}, eliminating potential noise from malformed tokens.

\paragraph{The dynamical emergence of rare token neurons.}

\begin{figure}[htbp]
  \centering
  \begin{minipage}{0.48\textwidth}
    \centering
    \includegraphics[width=\textwidth]{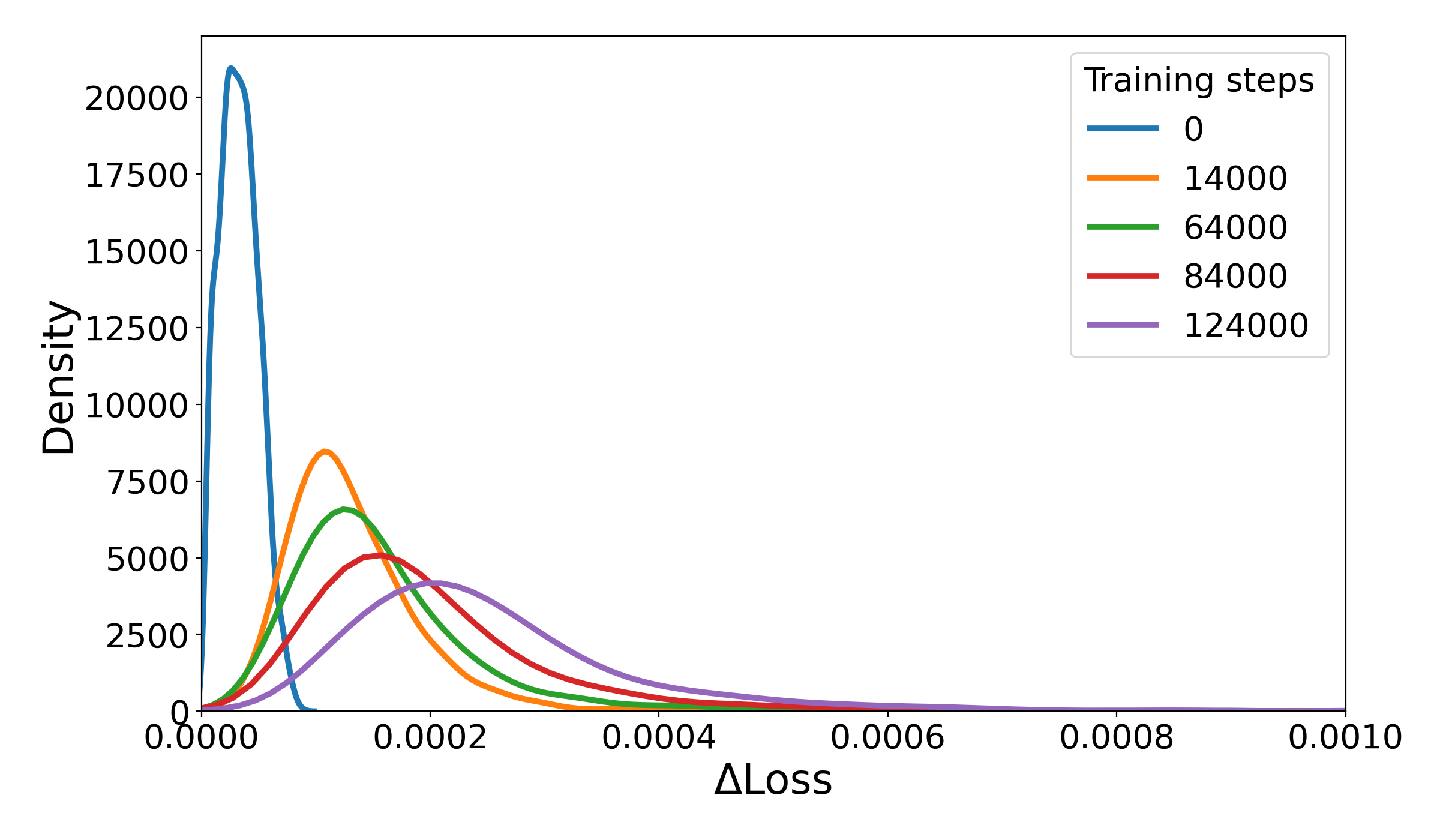}\\
    {\small (a) Absolute $\Delta$loss distribution in Pythia-410M model.}
  \end{minipage}
  \hfill
  \begin{minipage}{0.48\textwidth}
    \centering
    \includegraphics[width=\textwidth]{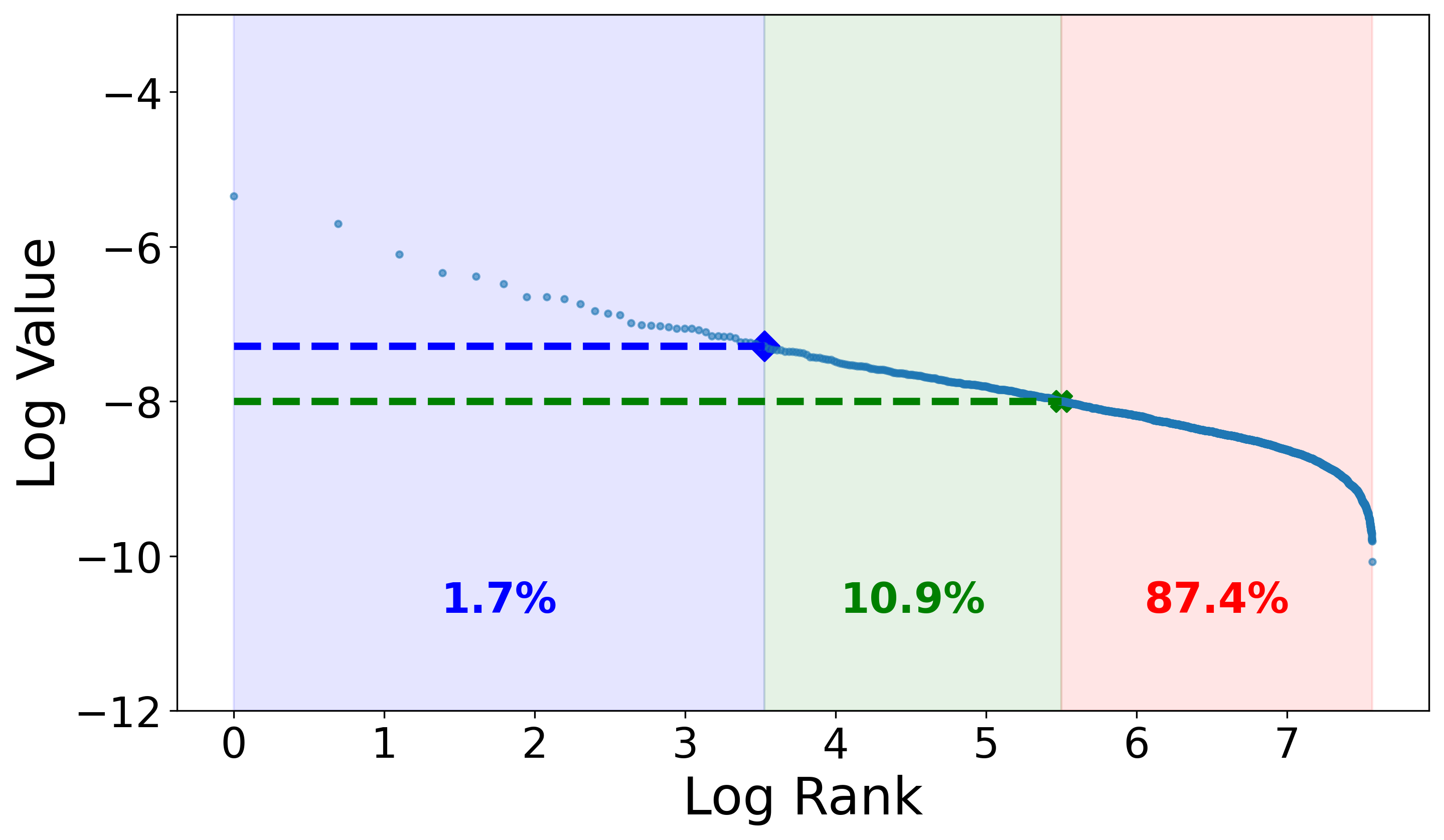}\\
    {\small (b) The three-phase structure of neuron influence.}
  \end{minipage}
  
  \caption{Neuron influence distribution and phase distinctions. The $\log\Delta$loss-$\log$rank relation shown in figure (b) reveals a three-phase structure: highly-influential plateau (blue) consisting of 1.7\% of neurons, mid-rank power-law phase (green) with 10.(\% of neurons, and low-rank rapid decay phase (red) with the remaining 87.4\% of neurons.}
  \label{fig:pythia_neuron_influence}
\end{figure}

Figure~\ref{fig:pythia_neuron_influence}a shows the distribution of per-neuron influence across training, measured by absolute change in token-level loss upon ablation. The condensation of neurons around zero-$\Delta$loss and a tail with large $\Delta$loss suggests that: through the training process, a small group neurons emerge particularly influential towards rare tokens, which we term the \textit{rare token neurons}. Within this group, we term neurons that boost (suppress) the appearance of rare tokens the boosting (suppressing) neurons.

\subsection{Distribution and Structure of Neuron Influence}\label{subsesction: Phases of neuron influence}

\paragraph{Identification of phases in neuron influence}

When we rank the neurons by their respective $\Delta\text{Loss}$, as computed in ~\ref{paragraph:delta loss def},  we observe a striking three-phase structure, presented on a log-log scale in (figure~\ref{fig:pythia_neuron_influence}b), which persists across model scales and architectures. Specifically, we observe the following phases:

\begin{enumerate}
    \item \textbf{Influential plateau phase:} A small fraction of neurons exhibit consistently high influence, forming a plateau on the leftmost of figure~\ref{fig:pythia_neuron_influence}b.
    
    \item \textbf{Power-law decay phase:} The majority of influential neurons follow a power-law relationship, which turns into a linear relation in log-log coordinates
    \begin{equation}\label{linear relation}
    \log |\Delta\text{Loss}| \approx -\kappa \log(\text{rank}) + \beta,
    \end{equation}
    where the power-law exponent $\kappa$ appears as the slope of a linear function. This aligns with theoretical predictions about sparse feature extraction in overparameterized networks~\citep{martin2021implicit}.
    
    \item \textbf{Rapid decay phase:} The influence of the remaining neurons on the rightmost of figure~\ref{fig:pythia_neuron_influence}b decays more rapidly than power-law predictions, indicating negligible contribution to rare token prediction.
\end{enumerate}

These phases  suggest computational specialization wherein a small subset of neurons assumes disproportionate responsibility for processing infrequent patterns. The power-law relationship in the intermediate regime is particularly significant, as it indicates scale-free organization characteristic of self-organized criticality in complex systems~\citep{bak1987self, watkins2016power}.

To precisely identify phase boundaries and track their evolutions during training, it is critical to understand the power-law exponent, appearing as a slope. We employ finite difference method with a sliding window for estimating this slope:

\begin{equation}
-\kappa(r) \approx -\frac{\log |\Delta\text{Loss}(r \cdot e)| - \log |\Delta f\text{Loss}(r)|}{\log(e)}
\end{equation}

where $r$ is the rank and $e$ is Euler's number. This finite-difference approximation provides a robust estimate of the local slope in log-log space, thus enabling the identification of the behavior of $-\kappa(r)$, in particular transition points where it changes significantly. The three phases are then identified using an automated change point detection algorithm~\citep{truong2020selective} applied to the $\kappa(r)$ curve, which identifies transition points where the slope changes dramatically. We validate these automatically detected boundaries through manual inspections for distribution differences on either side of the boundaries.

\subsection{Co-activation Patterns Through the Lens of Activation Space Geometry}

Having identified rare-token neurons through targeted ablation experiments, we turn to a mechanistic analysis of their behavior, aiming to uncover structural principles that govern the (dis)appearance of rare tokens in model predictions. To this end, we conduct a series of geometric analysis on the activation space. Our approach is motivated by the hypothesis that the internal representations learned by language models encode information in geometrically meaningful ways—such that certain geometric structures (e.g. vectors,subspaces or manifolds) are responsible for particular semantic representations \citep{park2024linearrepresentationhypothesisgeometry, park2025the}.

\paragraph{Construction of the activation space}


To understand how rare token neurons function collectively, we construct high-dimensional vectors comprising of activations of a certain neuron in response to the selected context-token pairs from the C4 corpus \citep{raffel2020exploring}.

\paragraph{Two geometric statistics for co-activation detection} 
We hypothesize that rare token neurons do not act in isolation, but instead participate in coordinated subspaces to modulate token-level probabilities. To this end we introduce two statistics in the activation space to measure the potential coordination patterns.

Firstly, we introduce the \emph{effective dimensionality} of each neuron's activation distribution using Principal Component Analysis (PCA). Formally, the effective dimension $d_\text{eff}$ is defined as the smallest $d$ such that the cumulative variance explained exceeds a fixed threshold $\tau$:
\[
d_\text{eff} = \min \left\{ d : \frac{\sum_{i=1}^{d} \lambda_i}{\sum_{j=1}^{N} \lambda_j} \geq \tau \right\},
\]
where $\lambda_i$ denotes the $i$-th eigenvalue of the activation covariance matrix.

The second statistics is the \textit{pairwise cosine similarity} between activation vectors, measuring the activation similarity between neurons, regardless of their activation intensities. Let $\mathbf{h}_i, \mathbf{h}_j \in \mathbb{R}^T$ denote activation traces across $T$ token contexts:
\[
\cos(\theta_{ij}) = \frac{\mathbf{h}_i \cdot \mathbf{h}_j}{\|\mathbf{h}_i\| \|\mathbf{h}_j\|}.
\]

\paragraph{Clustering and activation correlations} 
To investigate whether rare token neurons exhibit clustered activation patterns, we compute pairwise correlations of their activations across the selected context-token pairs. For each neuron pair $(i, j)$, we first calculate the Pearson correlation coefficient $\rho_{ij}$ between their activation vectors, then transform it into a distance metric:
\begin{equation}
D_{ij} = 1 - |\rho_{ij}|,
\end{equation}
which captures dissimilarity while remaining agnostic to the direction of correlation.

We apply hierarchical agglomerative clustering with Ward linkage to this distance matrix. Specifically, we measure the number of distinct clusters that emerge at a distance threshold of $t = 0.5$. A larger number of clusters would indicate greater functional modularity within the rare-token neuron population, while fewer clusters would suggest more globally coordinated behavior.

\subsection{Weight Eigenspectrum}\label{sec:Weight Eigenspectrum}

To understand the emergence of specialized neuron groups, we analyze model checkpoints across different training steps. This analysis enables us to track how the network progressively develops functional differentiation through the lens of Heavy-Tailed Self-Regularization (HT-SR) theory. 

HT-SR theory, introduced in Section \ref{sec:Heavy-Tailed Self-Regularization (HT-SR) Theory} suggests that heavy-tailed structures emerge from feature learning, where useful correlations are extracted during optimization. Neuron groups with more heavy-tailed ESDs which contain more learned signals, are assigned lower sparsity, while neuron groups with light-tailed ESDs are assigned higher sparsity. In practice, for each neuron group $\mathcal{G}$, we compute its correlation matrix as
\[
\mathbf{\Xi}_{\mathcal{G}} = \frac{1}{d} \mathbf{W}_{\mathcal{G}} \mathbf{W}_{\mathcal{G}}^\top,
\]
where $\mathbf{W}_{\mathcal{G}} \in \mathbb{R}^{|\mathcal{G}| \times d}$ denotes the slice of the weight matrix corresponding to the group $\mathcal{G}$. We then analyze the eigenvalue spectrum $\{\lambda_i\}$ of $\mathbf{\Xi}_{\mathcal{G}}$ to assess the internal dimensionality and structure of the group's learned representations.

To quantify the spectral shape, we use the Hill estimator to measure the power-law exponent $\alpha_{\text{Hill}}$ in the tail of the eigenvalue distribution:
\begin{equation}
\alpha_{\text{Hill}} = \left[ \frac{1}{k} \sum_{i=1}^{k} \log\left( \frac{\lambda_i}{\lambda_k} \right) \right]^{-1},
\end{equation}

where $k$ is a tunable parameter that adjusts the lower eigenvalue threshold $\lambda_{\text{min}}$ for (truncated) PL estimation. Following prior work on layer-wise pruning \citep{lu2024alphapruning}, we apply the Fix-finger method \citep{yang2023test} to select the $k$, which sets $k$ to align $\lambda_{\text{min}}$ with the peak of the ESD. By tracking the evolution of $\alpha_{\text{Hill}}$ across training, we can infer how specialized substructures or subnetworks progressively form and adapt.

\section{Results}

\subsection{Phases of Influence and Phase Transitions}


Our analysis reveals a three-phase structure of neuron influence that emerges dynamically during language model training. Here, we provide quantitative analysis of these phases to track their emergence during training.

\paragraph{The power-law and rapid decay phases}

The power-law phase is characterized as a $\log$-rank regime where $\log(\Delta \text{loss})$ follows the linear relation (\ref{linear relation}) with respect to 
$\log(\text{rank})$:
\begin{equation*}
    \log |\Delta\text{Loss}| \approx -\kappa \log(\text{rank}) + \beta.
\end{equation*}

\begin{figure}[htbp]
  \centering
  \begin{minipage}{0.48\textwidth}
    \centering
    \includegraphics[width=\textwidth]{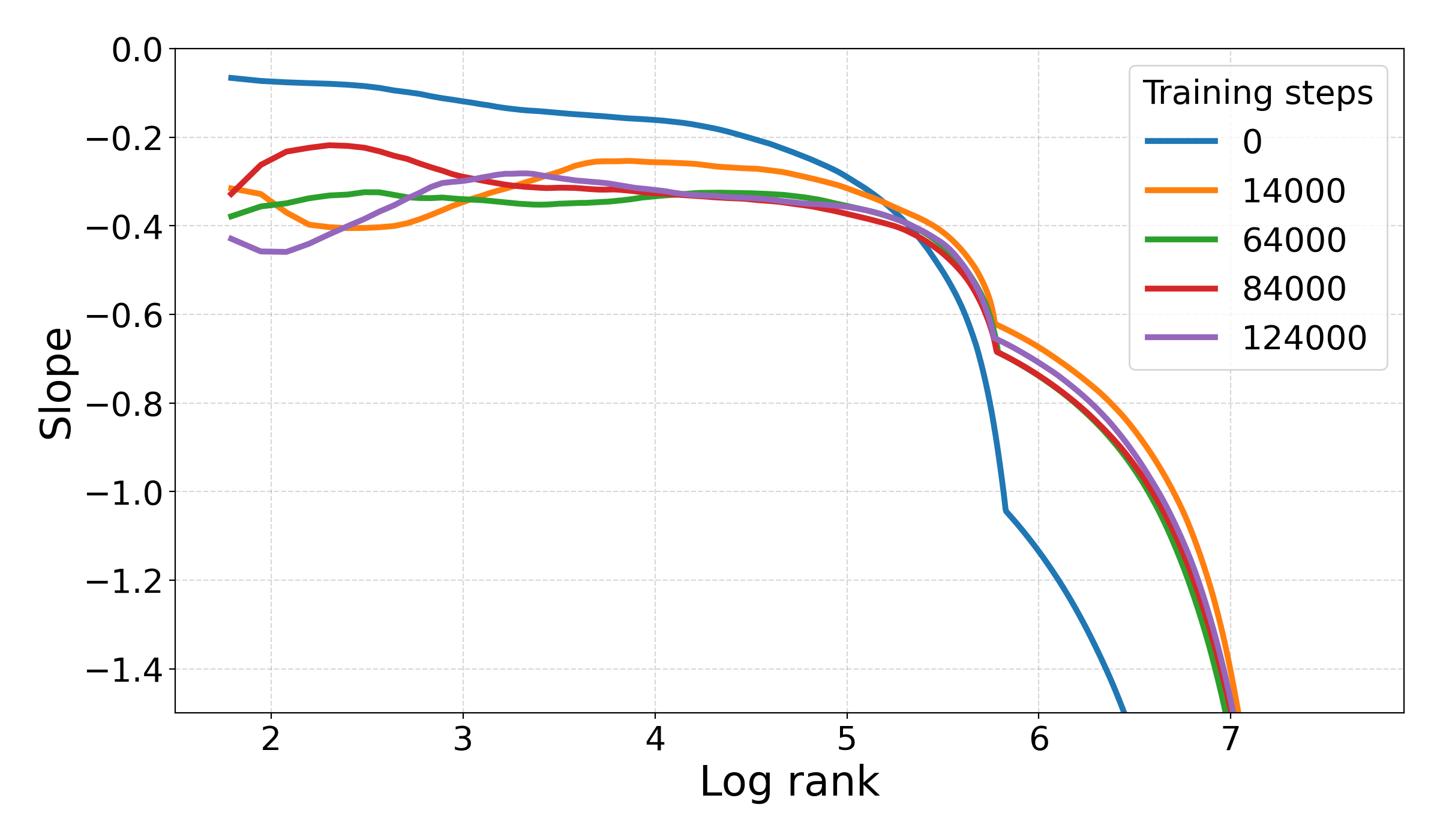}
    {\small (a) Neuron slope distributions.}
  \end{minipage}
  \hfill
  \begin{minipage}{0.48\textwidth}
    \centering
    \includegraphics[width=\textwidth]{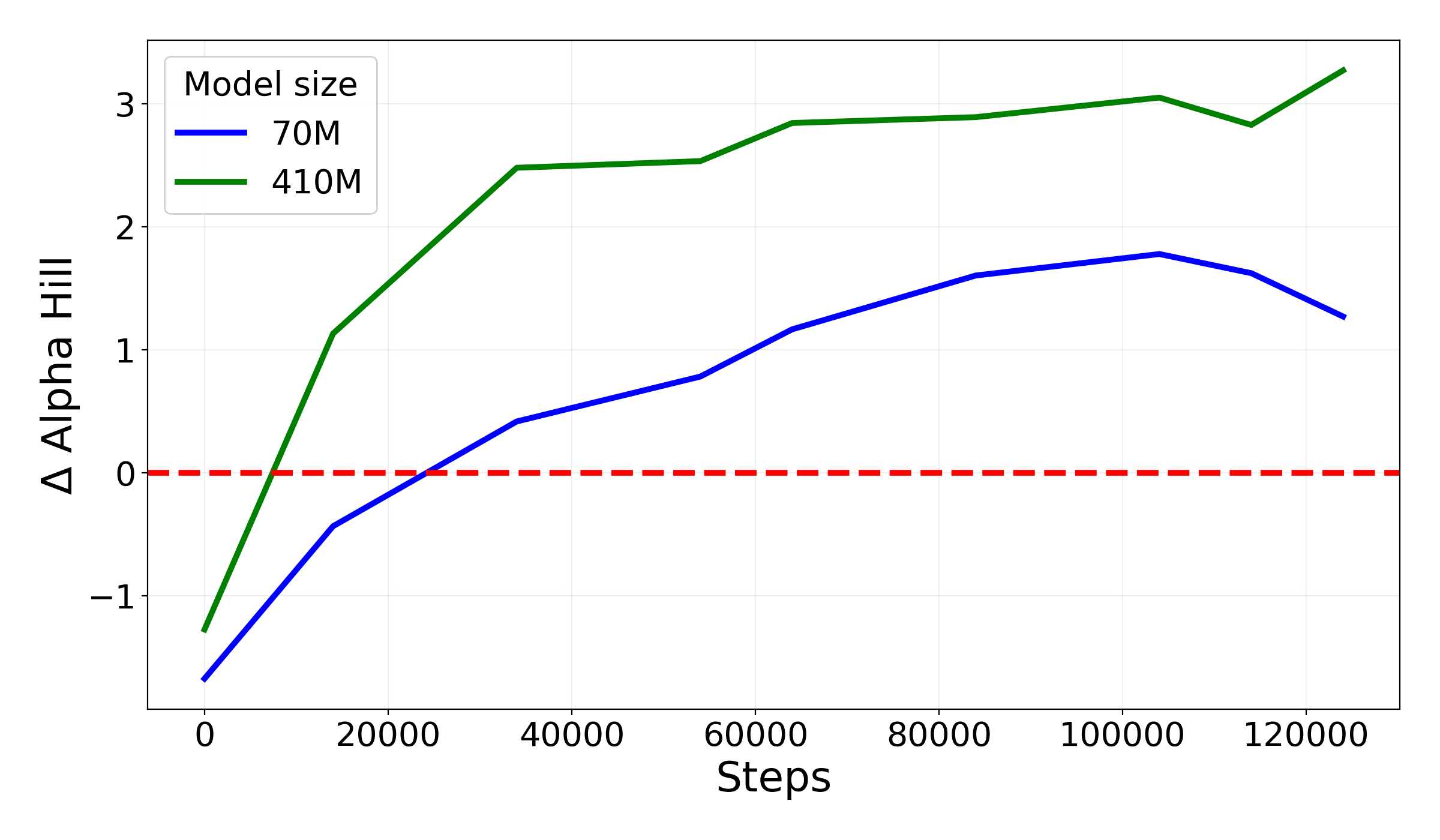}
    {\small (b) Difference in power-law exponents ($\alpha_{\text{Hill}}$).}
    
  \end{minipage}
  \caption{Parallel emergence of functional specialization and statistical heavy-tailedness during model training. (a) The slope distribution evolves to form distinct neuronal regimes at higher training steps. (b) Specialized neurons develop increasingly heavy-tailed weight distributions compared to random neurons, hinting a link between functional differentiation and statistical properties.}
  \label{fig: first derivative and htsr}
\end{figure}

As shown in figure~\ref{fig: first derivative and htsr}a, the first derivative of $\log(\Delta \text{loss})$ with respect to $\log(\text{rank})$ exhibits a quick drop around $\log\text{rank}=5$. This abrupt change marks the failure of power-law for least influential neurons, and characterizes the boundary between the power-law phase and the rapid decay phase.


\paragraph{Dynamical emergence of a highly-influential plateau}

\begin{figure}[htbp]
  \centering
  \begin{minipage}{0.48\textwidth}
    \centering
    \includegraphics[width=\textwidth]{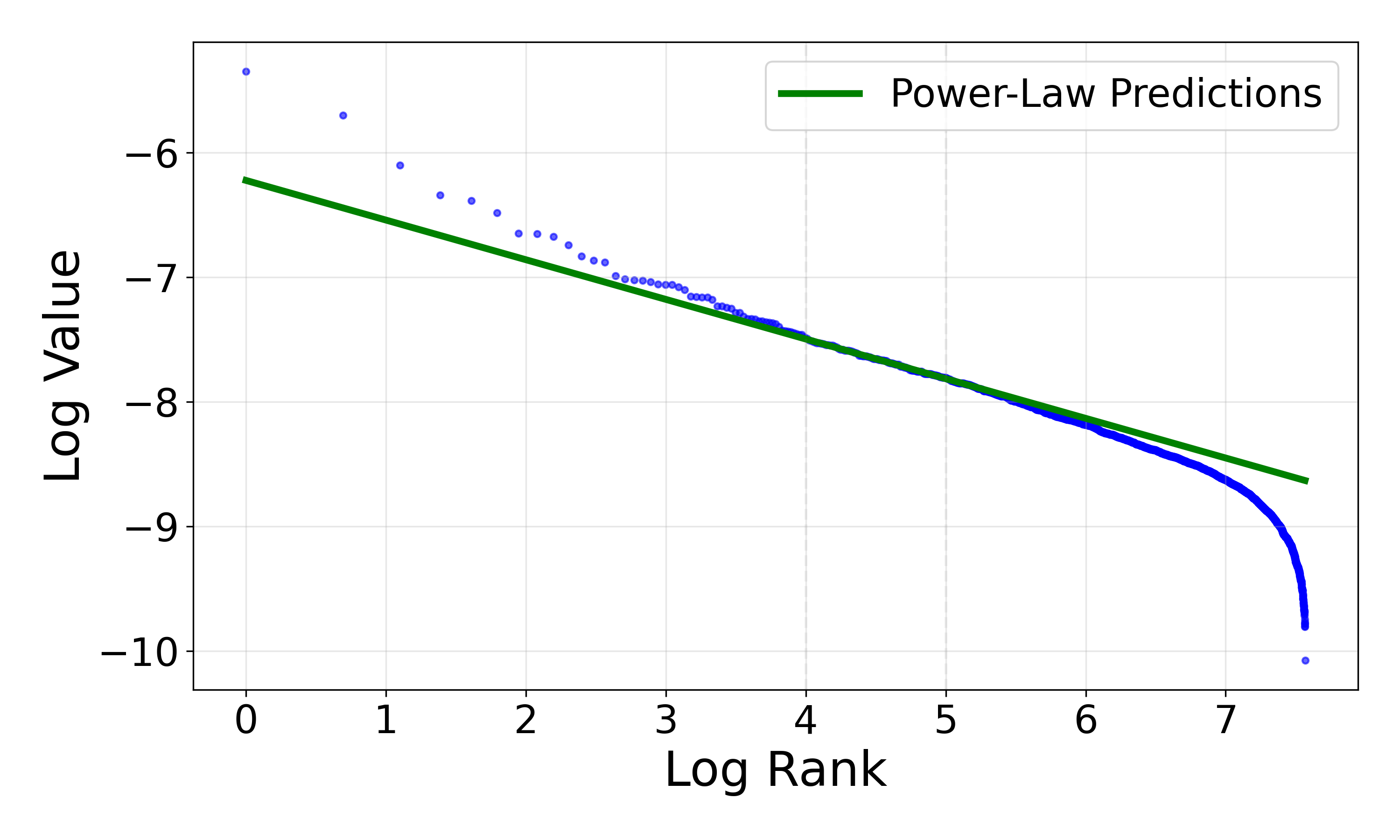}
    {\small (a) Power-law and its failure on both ends.}
  \end{minipage}
  \hfill
  \begin{minipage}{0.48\textwidth}
    \centering
    \includegraphics[width=\textwidth]{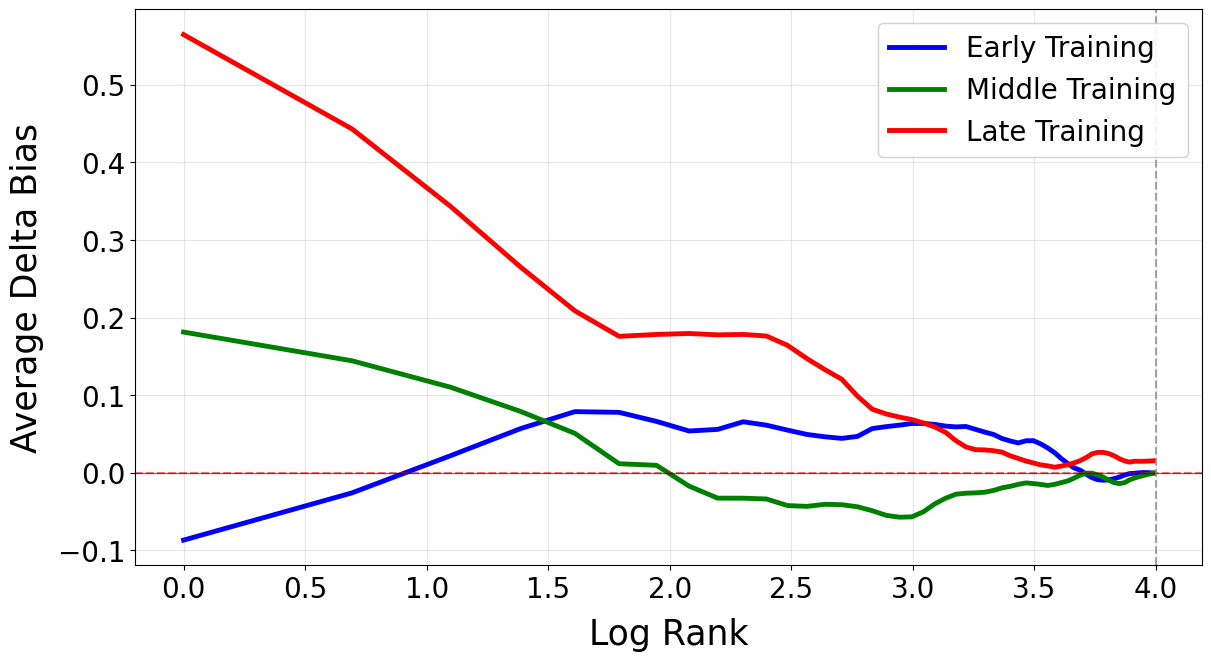}
    {\small (b) Emergence of plateau phase through additional bias term.}
  \end{minipage}
  
  \caption{As shown in (a), in $\log-\log$ coordinates, the green line indicates the power law. For the least influential neurons on the rightmost, power-law fails for a rapid drop of influence. While for the most influential neurons on the leftmost, the power-law fails due to an emergence of additional bias, while the slope remains as the power-law regime. In (b) we illustrate the dynamical deviation of top-ranked neurons from power-law predictions. At early steps of training (blue), the bias is close to 0 across log rank, indicating the power-law describes the whole rank regime $\log\text{rank}\in (0,4)$. While as training proceeds to later steps (green), top-ranked neurons deviate from power law prediction and from a plateau around the regime $\log\text{rank}\in (0,1.5)$. This plateau becomes more evident at late training steps (red), where neurons ranked $\log\text{rank}\in (0,3)$ all deviate from power-law prediction significantly.}
  \label{fig: power-law and its failure}
\end{figure}

Unlike the rapid decay phase, the first derivative fails to distinguish the plateau from the power-law phase. As in Figure~\ref{fig: first derivative and htsr}a, neurons in the range $\log(\text{rank})\in (2,5)$ exhibit an approximately uniform first derivative, indicating that the plateau keeps a power-law behavior. Yet, as in figure~\ref{fig: power-law and its failure}a, the highly-influential neurons systematically deviate from the power-law predictions.


We quantify this deviation by calculating the difference between observed influence values $\log|\Delta\text{Loss}(r)|$ and the power-law prediction $(-\kappa \log(r) + \beta)$:
\begin{equation}
\delta(r) = \log|\Delta\text{Loss}(r)| - (-\kappa \log(r) + \beta)
\end{equation}
where $\kappa$ and $\beta$ are parameters estimated from the power-law phase region. The quantity $\delta(r)$ illustrates how much the ranked neuron $r$ deviates from the power-law prediction. \textit{A plateau phase is therefore characterized as a $\log\text{rank}$ range where $\delta(r)$ is bounded above a positive value, hence the name ``plateau".}

In figure~\ref{fig: power-law and its failure}b, we illustrate the dynamics of $\delta(r)$ through the training process. It shows that \textit{the plateau phase emerges progressively during training}. The deviation is most pronounced for highest-ranked neurons and develops gradually as training proceeds, becoming increasingly significant in the later training stages. This evolution demonstrates a process of progressive functional differentiation, where a small subset of neurons gains disproportionate influence beyond what would be predicted by the power-law relationship.


Notably, these plateau-phase neurons maintain slope characteristics similar to those in the power-law phase but operate at a higher baseline level of influence. As training proceeds, they acquire an additional positive bias term that causes systematic deviation from power-law scaling. The temporal development of these phases indicates that language models progressively form a specialized neuron subnetwork for rare token processing.

\paragraph{A hidden singularity and possible second-order phase transition}

Our analysis reveals an additional phenomenon of theoretical interest at the boundary between the power-law and rapid decay phases.

As shown in Figure~\ref{fig: first derivative and htsr}a, we observe a sharp discontinuity in the derivative of the slope function. Specifically, while the first derivative of $\log|\Delta\text{Loss}|$ with respect to $\log(\text{rank})$ remains continuous, its rate of change (i.e., the second derivative) exhibits an apparent discontinuity. 

This mathematical signature is analogous to second-order phase transitions in statistical physics, where the first derivative of the free energy remains continuous while the second derivative exhibits a discontinuity. The presence of this singularity suggests that the transition between the power-law and rapid decay regimes may represent a genuine phase transition in the information-theoretic sense, rather than a mere change in scaling behavior. This observation provides empirical support for recent theoretical frameworks connecting neural network optimization to statistical mechanics \citep{bahri2020statistical}, where critical points in the loss landscape can induce structural reorganization of representational geometry. Furthermore, this phase transition boundary emerges progressively during training, becoming increasingly well-defined in later stage, which suggests that the critical phenomenon is an emergent property of the optimization process rather than an artifact of network initialization or architecture.


\subsection{Eigenspectral Specialization and Temporal Dynamics}

Figure~\ref{fig: first derivative and htsr}b illustrates how specialized neurons develop distinctly different statistical properties compared to random neurons across training. After the initial phase, $\alpha_{\text{Hill}}$ values for specialized neurons consistently shows lower values than those of random neurons, indicating heavier-tailed weight distributions in rare token neurons regardless of model sizes.

This persistent separation provides strong evidence for functional differentiation through implicit regularization. Despite fluctuations during training, the fundamental pattern remains: neurons that significantly impact rare token prediction consistently develop more pronounced heavy-tailed characteristics than neurons with random or general functionality.

These findings align with HT-SR theory, which posits that neural networks naturally organize towards criticality during optimization. The consistently lower $\alpha_{\text{Hill}}$ values in rare token neurons suggest that they operate closer to this critical regime—a state that optimizes the balance between stability and expressivity necessary for processing rare events in the training set.

The observed co-occurrence of functional specialization in neuron behavior and heavy-tailed weight distributions suggests a potential connection between these phenomena, though the exact relationship requires further investigation. We hypothesize that the heavy-tailed distribution framework provides the statistical foundation that enables certain neurons to exert disproportionate influence within the network. This statistical perspective offers a principled explanation for how neural networks develop the functional differentiation observed in our analysis—specifically the plateau, power-law, and rapid decay regimes—without explicit architectural constraints.


\subsection{Geometric Analysis in Activation Space}

The geometric analysis of neuron representations reveals a striking pattern of coordinated activity, that is, on the one hand, rare token neurons demonstrate significant co-activation with each other; and on the other hand, these neurons systematically \textbf{avoid} co-activation with neurons less responsible for rare token prediction. This emergent coordination is particularly notable, as our identification procedure considered only individual causal effects on rare token probabilities, without explicitly targeting activation correlations. 

The co-activation analysis reveal a strong activation correlation within the rare token neuron group while considerably low-to-zero correlation within the random baseline neurons. This spontaneous co-activation pattern suggests an intrinsic mechanism within th emodel rather than purely beyond the loss-based heuristics by probing.





\paragraph{Effective dimensionality}
Analysis on effective dimension proportion exhibits a significantly lower-dimensional manifold compared to randomly selected neurons (0.49 v.s. 0.56, t-test p < .05). This dimension compression indicates that rare token neurons occupy a more constrained manifold of the activation space and are likely to be activated in a coordinated manner rather than independently.

\paragraph{Activation alignment} 


Our pairwise cosine similarity analysis reveals distinct patterns of mechanisms within and scross neuron groups. Random neurons exhibit near-zero similarity($\overline{\cos\theta} \approx 0.03 \pm 0.04$), confirming their uncorrelated activation patterns. In contrast, neurons within both the boosting and suppressing rare token groups demonstrate substantial positive similarity scores ($\overline{\cos\theta} \approx 0.41 \pm 0.12$), reflecting functional specialization. Interestingly, despite their opposing effects on token probabilities, boosting and suppressing neurons maintain substantial positive correlation ($\overline{\cos\theta} \approx 0.32 \pm 0.09$), suggesting coordinated functional roles, indicating antagonistic activation patterns. 

These findings reveal a structured organization where rare token neurons operate in coordinated opposition to random neurons.





\section{Discussions and Conjectures}
Our analysis reveal distinct phases of neuronal influence that emerge through training. Within the influential plateau and the power-law phases we identify co-activation patterns and heavy-tailed spectral statistics. We summarize these empirical observations into two mechanistic conjectures:
 \begin{hypt}[Dual-Regime Organization]

     The emergence of power-law phase and its distinction from the rapid decay phase suggests a spontaneous specialization of influential neurons. Among this group of influential neurons, the power-law structure, the $\alpha_{\text{Hill}}$ behavior and the co-activation patterns are signs of self-organization phenomena.
 \end{hypt}

This conjecture is supported by the abrupt transition in slope between the power-law and rapid decay phases, suggesting a qualitative change in neuronal function rather than a continuous gradient. Within the power-law group, the heavy-tailed statistics and coordinated activation patterns imply a rich inner structure.

\begin{hypt}[Parallel Mechanism Conjecture]
    The plateau phase emerges through a mechanism that parallels the power-law mechanism. It respects the power-law mechanism, but differentiates a small subset of neurons within the power-law group by lifting their influence to form the influential plateau.
\end{hypt}

This conjecture is supported by three key observations: (1) the plateau phase maintains a similar local slope to the power-law phase, suggesting it respects the same underlying scaling principle despite their increased influence; (2) the plateau emerges progressively during training rather than being present from initialization; and (3) the magnitude of deviation from the power-law fit increases systematically with neuron rank within the plateau, indicating structured rather than random differentiation. 

The parallel mechanism conjecture suggests that rare token processing in language models involves both distributed computation (power-law phase) and specialized neuron group (plateau phase). This dual-system architecture resembles the complementary learning systems (CLS) observed in the human memory system \citep{mcclelland1995there, kumaran2016learning}, where general statistical learning occurs alongside specialized mechanisms for handling exceptional cases. Analogously, the plateau neurons might function as a specialized memory system for encoding rare linguistic patterns that would otherwise be overwhelmed by the statistics of other tokens.

\section{Conclusion}

This paper presents a systematic investigation into the emergent neuronal mechanisms that language models develop for processing rare tokens—a fundamental challenge requiring a balance between statistical efficiency and representational capacity for low-frequency events. Through targeted ablation experiments across a range of models, we identified a small subset of neurons with disproportionate influence on rare token prediction, and demonstrated that these neurons exhibit coordinated activation patterns, including significant co-activation among themselves and systematic anti-correlation with neurons processing common tokens.

Our temporal analysis revealed a three-phase structure of neuronal influence, consisting of a specialized influential plateau phase, a power-law phase following efficient coding principles, and a rapid decay phase with minimal contribution to rare token processing. We observed evidence of a phase transition between regimes and found that functionally specialized neurons develop more pronounced heavy-tailed weight distributions, suggesting operation closer to criticality.

Based on these findings, we proposed  \textit{the Dual-Regime Organization} conjecture, suggesting qualitatively different computational regimes across neuron groups, and \textit{the Parallel Mechanism} conjecture, positing that rare token processing involves both distributed and specialized computation analogous to complementary learning systems in biological memory.

This work represents the first comprehensive investigation into how language models develop mechanisms for rare token processing. Our findings demonstrate that these models spontaneously develop functionally specialized subnetworks—an emergent property that could inform future research for data-efficient model training and domain adaptations. Future research could explore whether similar principles govern other forms of specialization and scale with model size.

\newpage
\bibliography{main}

\newpage
\appendix
\section{Appendix}

\subsection{Limitations}

Our study presents several important limitations that warrant acknowledgment. First, our analysis focuses exclusively on neurons in the final MLP layer, while rare token processing likely involves distributed mechanisms across multiple architectural components throughout the model. Future work examining attention heads, intermediate layers, and cross-layer interactions could provide a more comprehensive understanding of rare token processing mechanisms.

Second, our methodology relies on ablation-based proxies, particularly change in loss, to quantify individual neuron contributions. While these measures provide useful insights, they lack the theoretical rigor of more principled attribution methods. Developing precise theoretical frameworks for measuring neuron-level effects would strengthen our mechanistic interpretations and enable more robust conclusions about functional specialization.

Finally, our analysis is constrained to next-token prediction in autoregressive language modeling contexts. The generalizability of our findings to downstream applications---including question-answering, mathematical reasoning, and domain-specific tasks---remains an open question. Investigating rare token processing mechanisms in these applied settings would illuminate their practical significance for real-world model performance.

\subsection{Results}

\paragraph{Activation Correlation Analysis}

We examined activation patterns across neuron groups in five pre-trained language models, comparing rare token neurons (boost and suppress groups) against random baseline controls. Table~\ref{tab:activation} presents pairwise activation correlations within and between neuron groups (group size = 50). 

The results demonstrate consistently higher intra-group correlations for both boost neurons (range: 0.004--0.036) and suppress neurons (range: 0.011--0.052) compared to random controls (range: -0.007--0.017). Notably, cross-group correlations between boost and suppress neurons (B vs. S) show positive values (0.010--0.040), suggesting coordinated but potentially antagonistic functionality. Random baseline comparisons (R1 vs. R2) exhibit near-zero correlations, confirming the specificity of our identified neuron groups.

\begin{table}[htbp]
\caption{Pairwise activation correlations within and between neuron groups}
\label{tab:activation}
\centering
\begin{tabular}{lcccccccc}
\toprule
\textbf{Model} & \textbf{Size} & \textbf{Boost} & \textbf{Suppress} & \textbf{Random} & \textbf{B vs. R} & \textbf{S vs. R} & \textbf{B vs. S} & \textbf{R1 vs. R2} \\
\midrule
Pythia-70M & 70M & 0.028 & 0.045 & 0.011 & 0.002 & 0.005 & 0.027 & 0.009 \\
Pythia-410M & 410M & 0.036 & 0.052 & 0.007 & 0.005 & 0.006 & 0.040 & 0.007 \\
GPT2-Small & 124M & 0.017 & 0.019 & 0.017 & -0.001 & -0.001 & 0.021 & 0.023 \\
GPT2-Large & 774M & 0.004 & 0.011 & 0.012 & -0.004 & -0.004 & 0.010 & 0.016 \\
GPT2-XL & 1.5B & 0.036 & 0.016 & -0.007 & 0.003 & -0.0004 & 0.020 & 0.008 \\
\bottomrule
\end{tabular}
\end{table}

\paragraph{Effective Dimensionality}

We computed the effective dimensionality of activation patterns using the participation ratio metric to assess functional specialization. Table~\ref{tab:dimensionality} shows that rare token neuron groups consistently exhibit lower effective dimensionality compared to random baselines. This reduction in effective dimensions (boost: 33.0--43.0\%, suppress: 32.2--43.0\%) relative to random controls (36.2--46.0\%) indicates more concentrated, specialized activation patterns within rare token neurons.

\begin{table}[htbp]
\caption{Effective dimensionality proportions across neuron groups}
\label{tab:dimensionality}
\centering
\begin{tabular}{lcccc}
\toprule
\textbf{Model} & \textbf{Size} & \textbf{Boost} & \textbf{Suppress} & \textbf{Random} \\
\midrule
Pythia-70M & 70M & 33.5 & 32.6 & 36.2 \\
Pythia-410M & 410M & 33.0 & 32.2 & 37.3 \\
GPT2-Small & 124M & 37.0 & 40.0 & 45.0 \\
GPT2-Large & 774M & 43.0 & 43.0 & 46.0 \\
GPT2-XL & 1.5B & 40.0 & 42.0 & 46.0 \\
\bottomrule
\end{tabular}
\end{table}

\paragraph{Cosine Similarity of Weight Vectors}

Weight vector cosine similarities provide insight into the geometric organization of rare token neurons in parameter space. Table~\ref{tab:cosine} reveals strong positive alignment within boost (0.028--0.141) and suppress (0.092--0.165) neuron groups, while showing consistent negative alignment between these specialized groups and random controls. The negative cross-correlations (boost vs. random: -0.100 to 0.003, suppress vs. random: -0.105 to -0.014) suggest that rare token neurons occupy distinct regions of weight space, supporting our hypothesis of functional specialization.

\begin{table}[htbp]
\caption{Cosine similarity between weight vectors within and across neuron groups}
\label{tab:cosine}
\centering
\begin{tabular}{lcccccccc}
\toprule
\textbf{Model} & \textbf{Size} & \textbf{Boost} & \textbf{Suppress} & \textbf{Random} & \textbf{B vs. R} & \textbf{S vs. R} & \textbf{B vs. S} & \textbf{R1 vs. R2} \\
\midrule
Pythia-70M & 70M & 0.141 & 0.165 & 0.021 & -0.017 & -0.014 & 0.146 & 0.021 \\
Pythia-410M & 410M & 0.107 & 0.133 & 0.054 & -0.032 & -0.041 & 0.114 & 0.058 \\
GPT2-Small & 124M & 0.109 & 0.122 & 0.089 & -0.100 & -0.105 & 0.120 & 0.099 \\
GPT2-Large & 774M & 0.028 & 0.092 & 0.041 & -0.034 & -0.063 & 0.054 & 0.052 \\
GPT2-XL & 1.5B & 0.095 & 0.095 & 0.009 & -0.010 & -0.015 & 0.090 & 0.012 \\
\bottomrule
\end{tabular}
\end{table}

\paragraph{Weight Distribution Analysis}

We analyzed the heavy-tailed properties of weight distributions using power-law exponents ($\alpha$) estimated via the Hill estimator. Table~\ref{tab:htsr} demonstrates that rare token neuron groups exhibit significantly lower $\alpha$ values compared to random controls, indicating heavier-tailed weight distributions. This finding supports our hypothesis that functional specialization correlates with the emergence of heavy-tailed statistical properties in neural networks.

\begin{table}[htbp]
\caption{Power-law exponents ($\alpha$) for weight distributions across neuron groups}
\label{tab:htsr}
\centering
\begin{tabular}{lcccc}
\toprule
\textbf{Model} & \textbf{Size} & \textbf{Boost} & \textbf{Suppress} & \textbf{Random} \\
\midrule
Pythia-70M & 70M & 4.30 & 3.97 & 6.37 \\
Pythia-410M & 410M & 3.80 & 3.43 & 7.56 \\
GPT2-Small & 124M & 2.12 & 1.57 & 6.74 \\
GPT2-Large & 774M & 3.30 & 1.84 & 8.31 \\
GPT2-XL & 1.5B & 2.01 & 1.68 & 9.33 \\
\bottomrule
\end{tabular}
\end{table}

\paragraph{Phase Transition Dynamics}

\begin{figure}[htbp]
  \centering
  \includegraphics[width=0.8\textwidth]{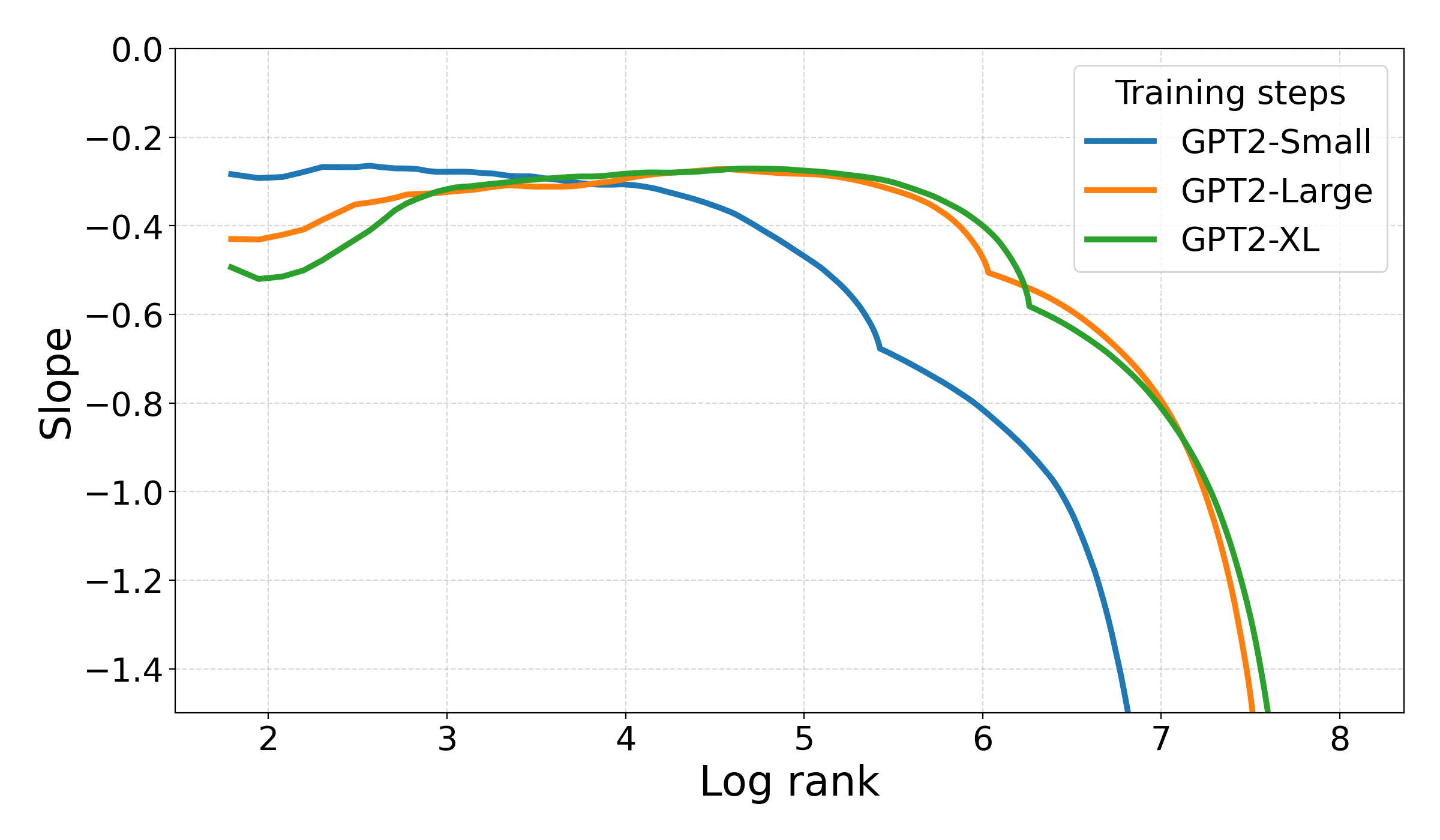}
  \caption{Distribution of neuron influence slopes for suppress neurons across the GPT-2 model family, showing characteristic three-phase organization emerging during training.}
  \label{fig:gpt_neuron_influence}
\end{figure}

Figure~\ref{fig:gpt_neuron_influence} illustrates the evolution of neuron influence distributions throughout training across the GPT-2 model family. The suppress neuron populations exhibit characteristic three-phase development: an initial plateau phase with uniform low influence, followed by a power-law scaling regime, and concluding with rapid decay at high influence values. This pattern emerges consistently across model scales, suggesting a universal mechanism underlying rare token neuron specialization.

\end{document}